\documentclass[11pt,a4paper]{article}
\usepackage{authblk}
\usepackage[hyperref]{acl2017}
\usepackage{times}
\usepackage{latexsym}

\usepackage{url}

\usepackage{microtype}
\usepackage{tikz}
\usepackage{nicefrac}
\usepackage[hang,flushmargin]{footmisc}
\usepackage{numprint}
\npthousandsep{,}\npthousandthpartsep{}\npdecimalsign{.}
\usepackage{multirow}

\aclfinalcopy 

\newcommand{\WikiQA}[0]{\textsc{WikiQA}}
\newcommand{\SelQA}[0]{\textsc{SelQA}}
\newcommand{\SQuAD}[0]{\textsc{SQuAD}}
\newcommand{\InfoQA}[0]{\textsc{InfoboxQA}}
\newcommand{\TAB}[0]{\:\;}

\title{Analysis of Wikipedia-based Corpora for Question Answering}

\author[1]{\bf Tomasz Jurczyk}
\author[2]{\bf Amit Deshmane}
\author[1]{\bf Jinho D. Choi}
\affil[1]{Mathematics and Computer Science, Emory University}
\affil[2]{Infosys Ltd.}
\affil[ ] {\textit {\{tomasz.jurczyk,jinho.choi\}@emory.edu}}
\affil[ ]{\textit {amit\_deshmane@infosys.com}}

\date{}

\begin{document}
\maketitle

\begin{abstract}
This paper gives comprehensive analyses of corpora based on Wikipedia for several tasks in question answering.
Four recent corpora are collected, \WikiQA, \SelQA, \SQuAD, and \InfoQA, and first analyzed intrinsically by contextual similarities, question types, and answer categories.
These corpora are then analyzed extrinsically by three question answering tasks, answer retrieval, selection, and triggering.
An indexing-based method for the creation of a silver-standard dataset for answer retrieval using the entire Wikipedia is also presented.
Our analysis shows the uniqueness of these corpora and suggests a better use of them for statistical question answering learning.
\end{abstract}

\section{Introduction}

Question answering (QA) has been a blooming research field for the last decade.
Selection-based QA implies a family of tasks that find answer contexts from large data given questions in natural language.
Three tasks have been proposed for selection-based QA.
Given a document, \textit{answer extraction}~\cite{shen2006exploring,sultan2016joint} finds answer phrases whereas \textit{answer selection}~\cite{wang:07a,yih2013question,yu:14a,wang:16a} and \textit{answer triggering}~\cite{yang:15a,jurczyk:16} find answer sentences instead, although the presence of the answer context is not assumed within the provided document for answer triggering but it is for the other two tasks.
Recently, various QA tasks that are not selection-based have been proposed~\cite{reddy2006dialogue,hosseini2014learning,jauhar-turney-hovy:2016:P16-1,sachan-dubey-xing:2016:P16-2}; however, selection-based QA remains still important because of its practical value to real applications (e.g., IBM Watson, MIT \textsc{Start}).

\noindent Several datasets have been released for selection-based QA.
\newcite{wang:07a} created the \textsc{QASent} dataset consisting of 277 questions, which has been widely used for benchmarking the answer selection task.
\newcite{feng:15a} presented \textsc{InsuranceQA} comprising 16K+ questions on insurance contexts.
\newcite{yang:15a} introduced \WikiQA\ for answer selection and triggering.
\newcite{jurczyk:16} created \SelQA\ for large real-scale answer triggering.
\newcite{rajpurkar2016squad} presented \SQuAD\ for answer extraction and selection as well as for reading comprehension.
Finally, \newcite{morales-EtAl:2016:EMNLP2016} provided \InfoQA\ for answer selection.

These corpora make it possible to evaluate the robustness of statistical question answering learning.
Although all of these corpora target on selection-based QA, they are designed for different purposes such that it is important to understand the nature of these corpora so a better use of them can be made.
In this paper, we make both intrinsic and extrinsic analyses of four latest corpora based on Wikipedia, \WikiQA, \SelQA, \SQuAD, and \InfoQA.
We first give a thorough intrinsic analysis regarding contextual similarities, question types, and answer categories (Section~\ref{sec:intrinsic-analysis}).
We then map questions in all corpora to the current version of English Wikipedia and benchmark another selection-based QA task, answer retrieval (Section~\ref{sec:answer-retrieval}).
Finally, we present an extrinsic analysis through a set of experiments cross-testing these corpora using a convolutional neural network architecture (Section~\ref{sec:extrinsic-analysis}).\footnote{All our resources are publicly available\\: \url{anonymous_url}}


\begin{table*}[htbp!]
\centering\small
\resizebox{\textwidth}{!}{
\begin{tabular}{c||c|c|c|c}
 & \multicolumn{1}{c|}{\textbf{\textsc{WikiQA}}} & \multicolumn{1}{c|}{\textbf{\textsc{SelQA}}} & \multicolumn{1}{c|}{\textbf{\textsc{SQuAD}}} & \multicolumn{1}{c}{\textbf{\textsc{InfoboxQA}}} \\
\hline\hline
Source & Bing search queries & Crowdsourced & Crowdsourced & Crowdsourced \\
Year   & 2015 & 2016 & 2016 & 2016 \\
(AE, AS, AT) & (O, O, O) & (X, O, O) & (O, O, X) & (X, O, X) \\
$(q,c,\nicefrac{c}{q})$ & $(\numprint{1242},\:\numprint{12153},\:9.79)$ & $(\numprint{7904},\:\numprint{95250},\:12.05)$ & $(\textbf{\numprint{98202}},\:\numprint{496167},\:5.05)$ & $(\numprint{15271},\:\numprint{271038},\:\textbf{17.75})$  \\
$(w, t)$ & $(\numprint{386440},\:\numprint{30191})$ & $(\numprint{3469015},\:\numprint{44099})$ & $(\numprint{19445863},\:\textbf{\numprint{115092}})$ & $(\numprint{5034625},\:\numprint{8323})$  \\
$(\mu_q, \mu_c$) & $(6.44,\:25.36)$ & $(11.11,\:25.31)$ & $(11.33,\:27.86)$ & $(9.35,\:9.22)$  \\
$(\Omega_q, \Omega_a, \Omega_{f})$ & $(46.72,\:\textbf{11.05},\:16.96)$ & $(32.79,\:16.98,\:20.19)$ & $(32.27,\:12.15,\:\textbf{16.54})$ & $(\textbf{26.80},\:35.70,\:28.09)$ \\
\end{tabular}}
\caption{\small Comparisons between the four corpora for answer selection. Note that both \WikiQA\ and \SelQA\ provide separate annotation for answer triggering, which is not shown in this table. The \SQuAD\ column shows statistics excluding the evaluation set, which is not publicly available. AE/AS/AT: annotation for answer extraction/selection/triggering, $q$/$c$: \# of questions/answer candidates, $w$/$t$: \# of tokens/token types, $\mu_{q/c}$: average length of questions/answer candidates, $\Omega_{q/a}$: macro average in \% of overlapping words between question-answer pairs normalized by the questions/answers lengths, $\Omega_{f}$: $\nicefrac{(2 \cdot \Omega_q \cdot \Omega_a)}{(\Omega_q + \Omega_a)}$.}
\label{tbl:intrinsic-analysis}
\vspace{-2ex}
\end{table*}

\section{Intrinsic Analysis}
\label{sec:intrinsic-analysis}

Four publicly available corpora are selected for our analysis.
These corpora are based on Wikipedia, so more comparable than the others, and have already been used for the evaluation of several QA systems.

\noindent\textbf{\WikiQA}~\cite{yang:15a} comprises questions selected from the Bing search queries, where user click data give the questions and their corresponding Wikipedia articles.
The abstracts of these articles are then extracted to create answer candidates.
The assumption is made that if many queries lead to the same article, it must contain the answer context; however, this assumption fails for some occasions, which makes this dataset more challenging.
Since the existence of answer contexts is not guaranteed in this task, it is called answer triggering instead of answer selection.

\vspace{1ex}\noindent\textbf{\SelQA}~\cite{jurczyk:16} is a product of five annotation tasks through crowdsourcing.
It consists of about 8K questions where a half of the questions are paraphrased from the other half, aiming to reduce contextual similarities between questions and answers.
Each question is associated with a section in Wikipedia where the answer context is guaranteed, and also with five sections selected from the entire Wikipedia where the selection is made by the Lucene search engine.
This second dataset does not assume the existence of the answer context, so can be used for the evaluation of answer triggering.

\vspace{1ex}\noindent\textbf{\SQuAD}~\cite{rajpurkar2016squad} presents 107K+ crowdsourced questions on 536 Wikipedia articles, where the answer contexts are guaranteed to exist within the provided paragraph.
It contains annotation of answer phrases as well as the pointers to the sentences including the answer phrases; thus, it can be used for both answer extraction and selection.
This corpus also provides human accuracy on those questions, setting up a reasonable upper bound for machines.
To avoid overfitting, the evaluation set is not publicly available although system outputs can be evaluated by their provided script.

\vspace{1ex}\noindent\textbf{\InfoQA}~\cite{morales-EtAl:2016:EMNLP2016} gives 15K+ questions based on the infoboxes from 150 articles in Wikipedia.
Each question is crowdsourced and associated with an infobox, where each line of the infobox is considered an answer candidate.
This corpus emphasizes the gravity of infoboxes, which summary arguably the most commonly asked information about those articles.
Although the nature of this corpus is different from the others, it can also be used to evaluate answer selection.

\begin{table*}[htbp!]
\centering\small
\begin{tabular}{c||c|c|c}
 & \multicolumn{1}{c|}{\textbf{\textsc{WikiQA}}} & \multicolumn{1}{c|}{\textbf{\textsc{SelQA}}} & \multicolumn{1}{c}{\textbf{\textsc{SQuAD}}} \\
\hline\hline
$(\rho, \gamma_c, \gamma_p), t \geq 0.3$ & $(\TAB 92.00,\:\numprint{1203},\:96.86)$ & $(90.00,\:\numprint{7446},\:94.28)$ & $(100.00,\:\numprint{93928},\:95.61)$ \\
$(\rho, \gamma_c, \gamma_p), t \geq \textbf{0.4}$ & $(\TAB \textbf{94.00},\:\textbf{\numprint{1139}},\:\textbf{91.71})$ & $(\textbf{94.00},\:\textbf{\numprint{7133}},\:\textbf{90.31})$ & $(\textbf{100.00},\:\textbf{\numprint{93928}},\:\textbf{95.61})$\\
$(\rho, \gamma_c, \gamma_p), t \geq 0.5$ & $(100.00,\:\numprint{1051},\:84.62)$    & $(98.00,\:\numprint{6870},\:86.98)$ & $(100.00,\:\numprint{93928},\:95.61)$\\
\hline\hline
$k = (1, \textbf{5}, 10, 20)$ & $(4.39, \textbf{12.47}, 16.59, 22.39)$ & $(20.01, \textbf{34.07}, 40.29, 46.40)$ & $(19.90, \textbf{35.08}, 40.96, 46.74)$ \\
\end{tabular}
\caption{\small Statistics of the silver-standard dataset (first three rows) and the accuracies of answer retrieval in \% (last row).\\$\rho$: robustness of the silver-standard in \%, $\gamma_{c/p}$: \#$/$\% of retrieved silver-standard passages (coverage).}
\label{tbl:mapping}
\vspace{-2ex}
\end{table*}

\subsection*{Analysis}

All corpora provide datasets/splits for answer selection, whereas only (\WikiQA, \SQuAD) and (\WikiQA, \SelQA) provide datasets for answer extraction and answer triggering, respectively.
\SQuAD\ is much larger in size although questions in this corpus are often paraphrased multiple times.
On the contrary, \SQuAD's average candidates per question ($\nicefrac{c}{q}$) is the smallest because \SQuAD\ extracts answer candidates from paragraphs whereas the others extract them from sections or infoboxes that consist of bigger contexts.
Although \InfoQA\ is larger than \WikiQA\ or \SelQA, the number of token types ($t$) in \InfoQA\ is smaller than those two, due to the repetitive nature of infoboxes.

All corpora show similar average answer candidate lengths ($\mu_c$), except for \InfoQA\ where each line in the infobox is considered a candidate.
\SelQA\ and \SQuAD\ show similar average question lengths ($\mu_q$) because of the similarity between their annotation schemes.
It is not surprising that \WikiQA's average question length is the smallest, considering their questions are taken from search queries.
\InfoQA's average question length is relatively small, due to the restricted information that can be asked from the infoboxes.
\InfoQA\ and \WikiQA\ show the least question-answer word overlaps over questions and answers ($\Omega_q$ and $\Omega_a$ in Table~\ref{tbl:intrinsic-analysis}), respectively.
In terms of the F1-score for overlapping words ($\Omega_f$), \SQuAD\ gives the least portion of overlaps between question-answer pairs although \WikiQA\ comes very close.

\begin{figure}[htbp!]
\centering
\includegraphics[scale=0.5]{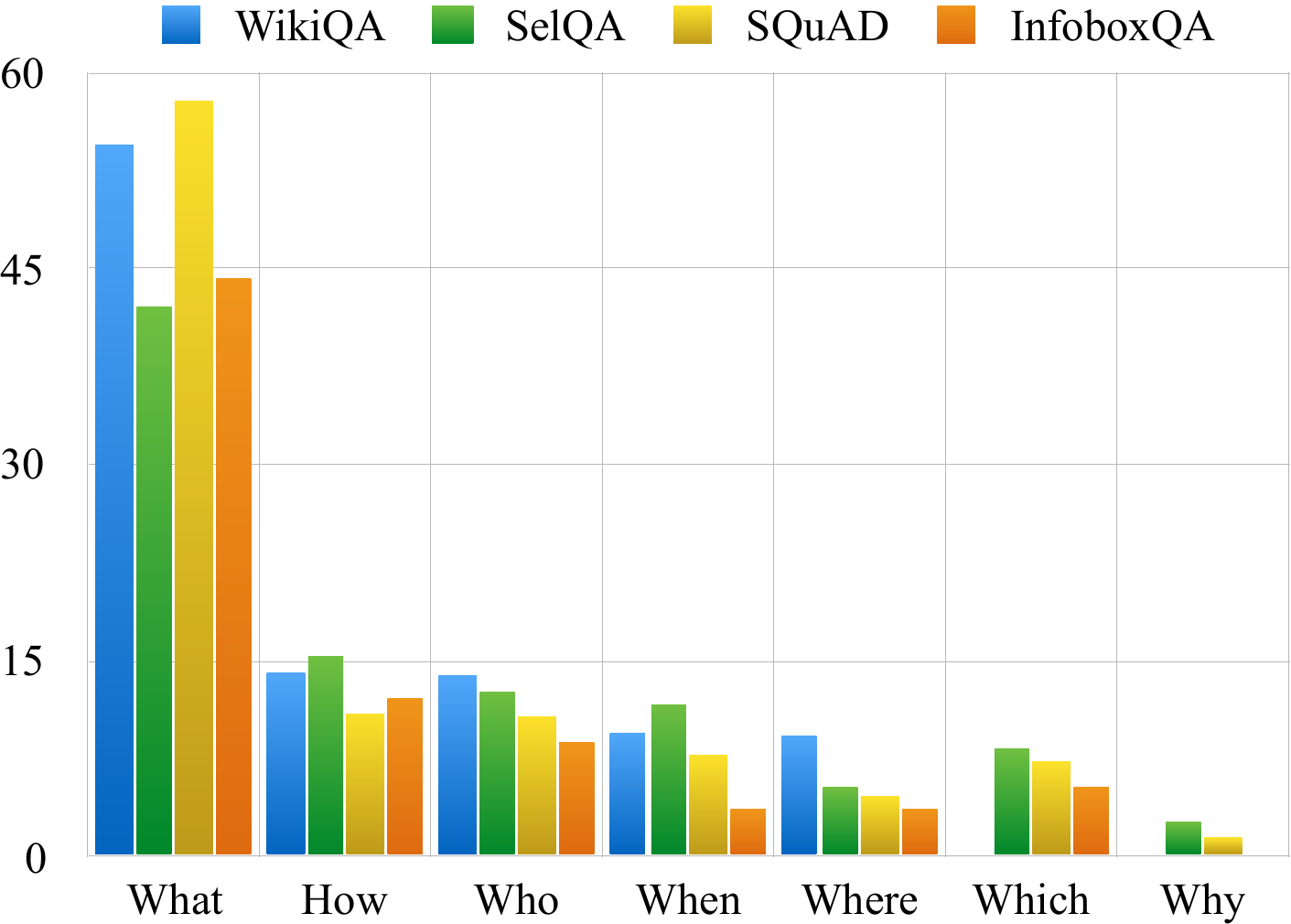}
\caption{Distributions of question types in \%.}
\label{fig:question-types}
\end{figure}

\noindent Fig.~\ref{fig:question-types} shows the distributions of seven question types grouped deterministically from the lexicons.
Although these corpora have been independently developed, a general trend is found, where the \textit{what} question type dominates, followed by \textit{how} and \textit{who}, followed by \textit{when} and \textit{where}, and so on.

\begin{figure}[htbp!]
\centering
\includegraphics[scale=0.5]{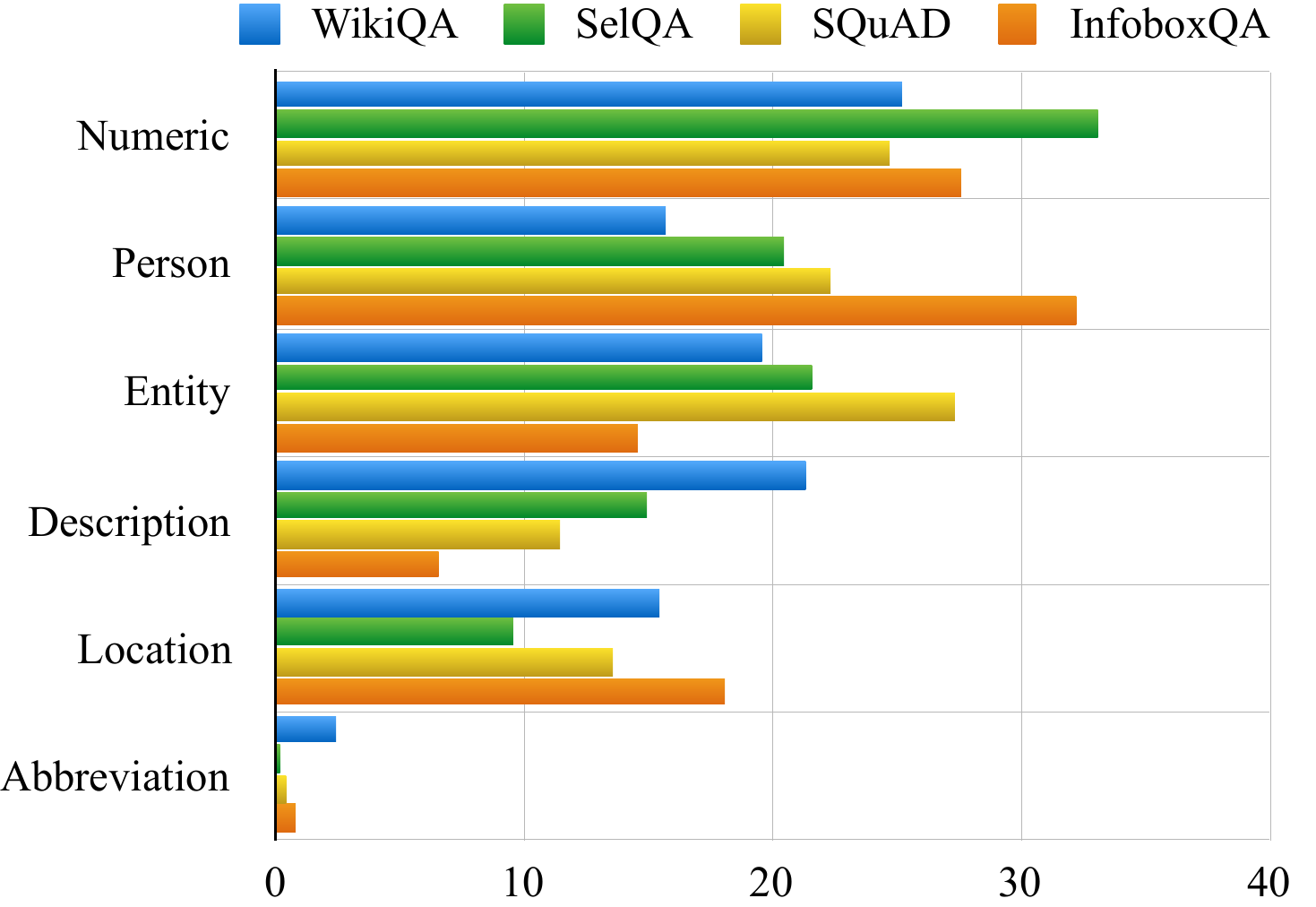}
\caption{Distributions of answer categories in \%.}
\label{fig:answer-categories}
\end{figure}

\noindent Fig.~\ref{fig:answer-categories} shows the distributions of answer categories automatically classified by our Convolutional Neural Network model trained on the data distributed by \newcite{li:02a}.\footnote{Our CNN model shows 95.20\% accuracy on their test set.}
Interestingly, each corpus focuses on different categories, \textit{Numeric} for \WikiQA\ and \SelQA, \textit{Entity} for \SQuAD, and \textit{Person} for \InfoQA, which gives enough diversities for statistical learning to build robust models.

\section{Answer Retrieval}
\label{sec:answer-retrieval}

This section describes another selection-based QA task, called \textit{answer retrieval}, that finds the answer context from a larger dataset, the entire Wikipedia.
\SQuAD\ provides no mapping of the answer contexts to Wikipedia, whereas \WikiQA\ and \SelQA\ provide mappings; however, their data do not come from the same version of Wikipedia.
We propose an automatic way of mapping the answer contexts from all corpora to the same version of Wikipeda\footnote{\url{enwiki-20160820-pages-articles.xml.bz2}} so they can be coherently used for answer retrieval.

Each paragraph in Wikipedia is first indexed by Lucene using \{1,2,3\}-grams, where the paragraphs are separated by WikiExtractor\footnote{\url{github.com/attardi/wikiextractor}} and segmented by NLP4J\footnote{\url{github.com/emorynlp/nlp4j}} (28.7M+ paragraphs are indexed).
Each answer sentence from the corpora in Table~\ref{tbl:mapping} is then queried to Lucene, and the top-5 ranked paragraphs are retrieved.
The cosine similarity between each sentence in these paragraphs and the answer sentence is measured for $n$-grams, say $n_{1,2,3}$.
A weight is assigned to each $n$-gram score, say $\lambda_{1,2,3}$, and the weighted sum is measured: $t = \sum_{i=1}^3 \lambda_i\cdot n_i$.
The fixed weights of $\lambda_{1,2,3} = (0.25, 0.35, 0.4)$ are used for our experiments, which can be improved.

If there exists a sentence whose $t \geq \theta$, the paragraph consisting of that sentence is considered the silver-standard answer passage.
Table~\ref{tbl:mapping} shows how robust these silver-standard passages are based on human judgement ($\rho$) and how many passages are collected ($\gamma$) for $\theta = [0.3, 0.5]$, where the human judgement is performed on 50 random samples for each case.
For answer retrieval, a dataset is created by $\theta = 0.4$, which gives $\rho \geq 94\%$ accuracy and $\gamma_p > 90\%$ coverage, respectively.\footnote{\SQuAD\ mapping was easier than the others because it was based on a more recent version of Wikipedia.}
Finally, each question is queried to Lucene and the top-$k$ paragraphs are retrieved from the entire Wikipedia.
If the answer sentence exists within those retrieved paragraphs according to the silver-standard, it is considered correct.


\begin{table*}[htbp!]
\centering\small
\resizebox{\textwidth}{!}{
\begin{tabular}{c||cc|c||cc|c||cc|c||cc|c}
\multirow{3}{*}{\bf Trained on} & \multicolumn{12}{c}{\bf Evaluated on} \\
 & \multicolumn{3}{c||}{\WikiQA} & \multicolumn{3}{c||}{\SelQA} & \multicolumn{3}{c||}{\SQuAD} & \multicolumn{3}{c}{\InfoQA} \\
 & MAP & MRR & F1 & MAP & MRR & F1 & MAP & MRR & F1 & MAP & MRR & F1 \\
\hline\hline
\WikiQA & \textbf{65.54} & \textbf{67.41} &         13.33  &         53.47  &         54.12  & $\TAB$   8.68  &         73.16  &         73.72  &         11.26  &         30.85  &         30.85  & - \\
\SelQA  &         49.05  &         49.64  & \textbf{24.30} &         82.72  &         83.70  & \textbf{48.66} &         77.22  &         78.04  &         44.70  &         63.13  &         63.13  & - \\
\SQuAD  &         58.17  &         58.53  &         19.35  &         81.15  &         82.27  &         42.88  &         88.84  &         89.69  & \textbf{44.93} &         63.24  &         63.24  & - \\
\InfoQA &         45.17  &         45.43  & -              &         53.48  &         54.25  & -              &         65.27  &         65.90  & -              & \textbf{79.44} & \textbf{79.44} & - \\
W+S+Q   &         56.40  &         56.51  & -              & \textbf{83.19} & \textbf{84.25} & -              &         88.78  &         89.65  & -              &         62.53  &         62.53  & - \\
W+S+Q+I &         60.19  &         60.68  & -              &         82.88  &         83.97  & -              & \textbf{88.92} & \textbf{89.79} & -              &         70.81  &         70.81  & - \\
\end{tabular}}
\caption{\small Results for answer selection and triggering in \% trained and evaluated across all corpora splits. The first column shows the training source, and the other columns show the evaluation sources. W: \WikiQA, S: \SelQA, Q: \SQuAD, I: \InfoQA.}
\label{tbl:extrinsic-analysis}
\end{table*}

\section{Extrinsic Analysis}
\label{sec:extrinsic-analysis}

\subsection{Answer Selection}

Answer selection is evaluated by two metrics, mean average precision (MAP) and mean reciprocal rank (MRR).
The bigram CNN introduced by \newcite{yu:14a} is used to generate all the results in Table~\ref{tbl:extrinsic-analysis}, where models are trained on either single or combined datasets.
Clearly, the questions in \WikiQA\ are the most challenging, and adding more training data from the other corpora hurts accuracy due to the uniqueness of query-based questions in this corpus.
The best model is achieved by training on W+S+Q for \SelQA; adding \InfoQA\ hurts accuracy for \SelQA\ although it gives a marginal gain for \SQuAD.
Just like \WikiQA, \InfoQA\ performs the best when it is trained on only itself.
From our analysis, we suggest that to use models trained on \WikiQA\ and \InfoQA\ for short query-like questions, whereas to use ones trained on \SelQA\ and \SQuAD\ for long natural questions.

\subsection{Answer Retrieval}
\label{ssec:experiment-answer-retrieval}
Finding a paragraph that includes the answer context out of the entire Wikipedia is an extremely difficult task (\nicefrac{1}{28.7M}).
The last row of Table~\ref{tbl:mapping} shows results from answer retrieval.
Given $k = 5$, \SelQA\ and \SQuAD\ show about 34\% and 35\% accuracy, which are reasonable.
However, \WikiQA\ shows a significantly lower accuracy of 12.47\%; this is because the questions in \WikiQA\ is about twice shorter than the questions in the other corpora such that not enough lexicons can be extracted from these questions for the Lucene search.

\subsection{Answer Triggering}

The results of $k = 5$ from the answer retrieval task in Section~\ref{ssec:experiment-answer-retrieval} are used to create the datasets for answer triggering, where about 65\% of the questions are not expected to find their answer contexts from the provided paragraphs for \SelQA\ and \SQuAD\ and 87.5\% are not expected for \WikiQA.
Answer triggering is evaluated by the F1 scores as presented in Table~\ref{tbl:extrinsic-analysis}, where three corpora are cross validated.
The results on \WikiQA\ are pretty low as expected from the poor accuracy on the answer retrieval task.
Training on \SelQA\ gives the best models for both \WikiQA\ and \SelQA.
Training on \SQuAD\ gives the best model for \SQuAD\ although the model trained on \SelQA\ is comparable.
Since the answer triggering datasets are about 5 times larger than the answer selection datasets, it is computationally too expensive to combine all data for training.
We plan to find a strong machine to perform this experiment in near future.

\section{Related work}

Lately, several deep learning approaches have been proposed for question answering.
\newcite{yu:14a} presented a CNN model that recognizes the semantic similarity between two sentences.
\newcite{wang-nyberg:2015:ACL-IJCNLP} presented a stacked bidirectional LSTM approach to read words in sequence, then outputs their similarity scores.
\newcite{feng:15a} applied a general deep learning framework to non-factoid question answering.
\newcite{santos:16a} introduced an attentive pooling mechanism that led to further improvements in selection-based QA.


\section{Conclusion}

We present a comprehensive comparison study of the existing corpora for selection-based question answering.
Our intrinsic analysis provides a better understanding of the uniqueness or similarity between these corpora.
Our extrinsic analysis shows the strength or weakness of combining these corpora together for statistical learning.
Additionally, we create a silver-standard dataset for answer retrieval and triggering, which will be publicly available.
In the future, we will explore different ways of improving the quality of our silver-standard datasets by fine-tuning the hyper-parameters.

\bibliography{acl2017}
\bibliographystyle{acl_natbib}

\end{document}